\documentclass{article}

\usepackage[final]{neurips_2024}

\usepackage[utf8]{inputenc} 
\usepackage[T1]{fontenc}    
\usepackage{hyperref}       
\usepackage{url}            
\usepackage{booktabs}       
\usepackage{amsfonts}       
\usepackage{nicefrac}       
\usepackage{microtype}      
\usepackage{xcolor}         

\usepackage{amsmath}
\usepackage{amsfonts}
\usepackage{algorithm}
\usepackage{algorithmic}
\usepackage{bm}
\usepackage{graphicx}
\usepackage{tcolorbox}
\tcbuselibrary{skins, breakable}
\usepackage{listings}
\usepackage{wrapfig}
\usepackage{caption}
\usepackage{float}
\usepackage{cleveref}
\usepackage{setspace}
\usepackage{afterpage}

\lstdefinestyle{python}{
    language=Python,
    basicstyle=\ttfamily\footnotesize,
    keywordstyle=\color{blue},
    commentstyle=\color{green!40!black},
    stringstyle=\color{red},
    stepnumber=1,
    numbersep=5pt,
    backgroundcolor=\color{gray!10},
    showspaces=false,
    showstringspaces=false,
    showtabs=false,
    tabsize=4,
    breaklines=true,
    breakatwhitespace=true,
    captionpos=b,
    keepspaces=true,
    columns=flexible,
}

\title{From Laws to Motivation: Guiding Exploration through Law-Based Reasoning and Rewards}

\author{%
  Ziyu Chen\footnotemark[1] \hspace{1cm}
  Zhiqing Xiao\footnotemark[1] \hspace{1cm}
  Xinbei Jiang \hspace{1cm}
  Junbo Zhao \\
  Zhejiang University\\
  \texttt{\{ziyu, zhiqing.xiao, xinbei, j.zhao\}@zju.edu.cn} \\
}

\begin{document}

\renewcommand{\thefootnote}{\fnsymbol{footnote}}
\footnotetext[1]{Equal contribution.}

\maketitle

\begin{abstract}
Large Language Models (LLMs) and Reinforcement Learning (RL) are two powerful approaches for building autonomous agents. However, due to limited understanding of the game environment, agents often resort to inefficient exploration and trial-and-error, struggling to develop long-term strategies or make decisions. We propose a method that extracts experience from interaction records to model the underlying laws of the game environment, using these experience as internal motivation to guide agents. These experience, expressed in language, are highly flexible and can either assist agents in reasoning directly or be transformed into rewards for guiding training. Our evaluation results in \texttt{Crafter} demonstrate that both RL and LLM agents benefit from these experience, leading to improved overall performance.
\end{abstract}

\section{Introduction}
Agents are often objective-driven. For agents based on Large Language Models (LLM), a well-defined objective makes them more efficient and effective at completing tasks~\cite{xi2023rise}. For Reinforcement Learning (RL) agents, objectives are expressed through reward, which shape the agent’s policy~\cite{dayan2002reward}. In simple environments, objectives may be singular, but in more complex environments, such as open-ended worlds, defining clear objectives becomes more challenging. In these environments, tasks can be subdivided into sub-tasks or involve multiple, sometimes conflicting, objectives, requiring agents to make trade-offs~\cite{dulac2019challenges}. These challenges often hinder agents’ performance.

However, humans, even in completely unfamiliar environments, can achieve strong performance through exploration. We believe that agents struggle to generalize well in complex, dynamic environments because they often lack a deep understanding of the underlying laws, properties, or mechanisms of environments. These laws, similar to physical laws in the real world, represent key characteristics of the environment. 

Humans are able to learn from brief interactions with the environment, leveraging accumulated commonsense to prune their action space and avoid unreasonable behaviors~\cite{schutz1962common}, while reflecting on past experience to form an understanding of environmental laws~\cite{boud2013reflection}. LLM agents retain context through memory modules to assist in reasoning~\cite{park2023generative, wang2023voyager, zhu2023ghost}, while RL agents rely on trials and feedback to evaluate actions~\cite{griffith2013policy, li2017deep}. 

We aim to leverage LLMs to infer experience from interaction records, approximate the laws of the environment, and use these "approximate laws" as internal motivation sources to guide agents' exploration. Experience is tied to actions. LLMs infer experience of an action by comparing the states before and after it is executed across multiple records (including failures and successes). For LLM agents, laws can be incorporated as context to improve reasoning and reduce hallucinations, while for RL agents, we propose the design of law-based rewards to promote more rational behaviors. These rewards, self-assigned by the agents, are independent of environmental feedback. Our goal is to enable agents to make decisions based on laws learned through interactions, allowing them to evaluate action outcomes before execution, relying on self-motivation rather than external feedback.

To demonstrate the benefits of understanding laws, we study agents in \texttt{Crafter}~\cite{hafner2021benchmarking}, a complex, open-ended environment similar to \texttt{Minecraft}~\cite{fan2022minedojo}. Our results show that LLMs can infer approximate laws of the environment from interactions, helping agents make more informed decisions. Our approach shifts agents from “blind exploration” to “purposeful exploration”, demonstrating the feasibility of using laws as internal motivation to guide exploration.

\begin{figure*}[t]
\vspace*{-3ex}
\centering
\resizebox{1\linewidth}{!}{\includegraphics{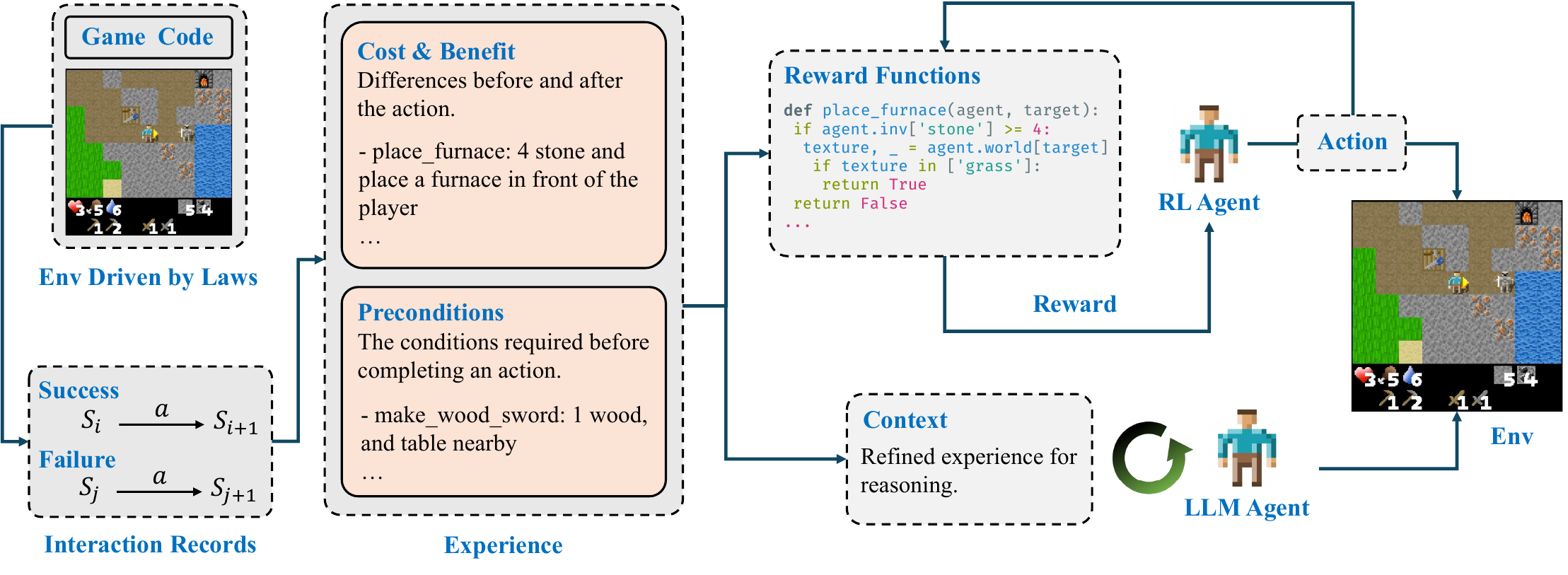}}
\caption{From Laws to Motivation. Experience approximates the laws of the environment and, through textual or reward, encourages the agents to achieve self-motivation.
}
\label{fig:overview}
\end{figure*}

\section{Related Works}
Related works are discussed in more detail in Appendix \ref{appendix:related}. With the recent advances of language models, more and more research attempts to incorporate external information into the design of agents~\cite{xi2023rise, zhu2023ghost}. When building LLM-based agents, static or dynamic environment-related information is introduced through in-context learning~\cite{sun2024adaplanner,  wang2023voyager, wu2024spring} or fine-tuning~\cite{wang2023adapting, xiang2024language}, enabling agents to make more adaptive decisions. Similarly, in the training of RL agents, external prior knowledge is expected to reduce trial-and-error~\cite{wu2024read}. One of the most common ways to leverage prior knowledge is through reward shaping~\cite{ng1999policy}, which reduces random exploration by providing auxiliary reward. Compared to traditional manual reward design, LLMs can generate code for reward functions based on the information provided in prompts~\cite{ma2023eureka, xie2023text2reward} or directly infer appropriate reward values~\cite{kwon2023reward}, significantly lowering the barrier from prior knowledge to reward. LLMs can also provide guidance to agents~\cite{du2023guiding, zhang2023omni}, thus helping agents in better learning.

\section{Method}
In this section, we introduce the specific methodology. First, we formulate the problem using a Constrained Markov Decision Process (CMDP)~\cite{altman2021constrained}. Then, we collect human player records and utilize LLMs to extract experience, denoted as \(\mathcal{E}\), from these records as an approximation of the laws. Records, similar to pre-collected data in Offline RL, focus on the description of states. The observation of state changes before and after actions, along with the effects of the same action under different states, reflects the laws of the game. Examples of records are provided in Appendix \ref{appendix:interactionlb}.
\(\mathcal{E}\) is in natural language form, describing the preconditions required to achieve each objective and the effects of achieving it, based on the perspective provided by the records.

\subsection{Constrained Markov Decision Process}
A CMDP is represented as $ \langle \mathcal{S}, \mathcal{A}, \mathcal{R}, \mathcal{P}, \gamma, \mathcal{C} \rangle $, which extends a standard MDP by incorporating additional constraints, denoted as $ \mathcal{C} $. The constraints $ \mathcal{C} = \{(c_i, b_i)\}_{i=1}^{m} $ consist of pairs, where each pair includes a cost function $ c_i $ and a corresponding threshold $ b_i $.

The cost function $ c_i(s, a) $ represents the cost incurred when action $ a $ is taken in state $ s $:
\begin{equation}
c_i: \mathcal{S} \times \mathcal{A} \rightarrow \mathbb{R},
\end{equation}
where $ \mathcal{S} $ is the state space and $ \mathcal{A} $ is the action space. Each constraint $ i $ in $ \mathcal{C} $ can be expressed as:
\begin{equation}
c_i(s, a) \leq b_i \quad \forall i \in \{1, \ldots, m\}.
\end{equation}

In the setting of objective-conditioned constraints, $ c_i = \mathbf{1}_{\{\cdot\}} $ is the indicator function that returns $ 1 $ if the condition inside the braces is met, and $ 0 $ otherwise. For all $ i $, $ b_i = 0 $.

Given any objective $ g_i $ associated with action $ a_{g_i} $, in a state $ s $, The transition to a new state $ s' $ is governed by the following conditional probabilities:
\begin{equation}
s' \sim 
\begin{cases} 
\mathcal{P}(\cdot \mid s, a_{g_i}) & \text{if } \bigwedge_{i=1}^m \left( c_i(s, a_{g_i}) \leq b_i \right) \\
\mathcal{P}(\cdot \mid s, \text{noop}) & \text{otherwise}
\end{cases}
\end{equation}
The achievement of objective $ g_i $ is defined as:
\begin{equation}
g_i \text{ is achieved} \iff \bigwedge_{i=1}^m \left( c_i(s, a_{g_i}) \leq b_i \right).
\label{f4_cmdp}
\end{equation}

\subsection{Generate $\mathcal{E}$}
We propose leveraging LLMs to extract and preserve essential constraints $\mathcal{C}$ from successful records $\mathcal{E}$ to guide agents in achieving objectives under specific conditions. More details and definitions can be found in Appendix \ref{appendix:records}.

Considering a record \( d = \langle s, a, s', v \rangle \). According to the definition of \( v \) and Equation \ref{f4_cmdp}, for any record \( d_k = \langle s_k, a_k, s'_k, v_k \rangle \), if \( a_k \) corresponds to the objective \( g_k \), then:
\begin{equation}
v_k = \text{True} \iff \bigwedge_{i=1}^m \left( c_i(s_k, a_k) \leq b_i \right).
\end{equation}

Let \( \mathcal{D} \) be the set of all records. Define \( \mathcal{D}_g = \{ d \in \mathcal{D} \mid a \text{ corresponds to } g \} \) as the subset of records in \( \mathcal{D} \). Further, define \( \mathcal{D}_g^{v=\text{True}} = \{ d \in \mathcal{D}_g \mid v = \text{True} \} \) and \( \mathcal{D}_g^{v=\text{False}} = \{ d \in \mathcal{D}_g \mid v = \text{False} \} \) as the subsets representing successful and unsuccessful attempts at objective \( g \), respectively.

\( \mathcal{D}_g^{v=\text{True}} \) is a good starting point because for all \( d = \langle s, a, s', v \rangle \in \mathcal{D}_g^{v=\text{True}} \), \( a \) is indicative of achieving \( g \) and \( s \) well reflects all necessary and unnecessary preconditions for \( g \). The state \( s \) is described in text. For all \( d = \langle s, a, s', v\rangle \in \mathcal{D}_g^{v=\text{True}} \), by comparing \( s \) and \( s' \), we can determine the cost and benefit of achieving the objective \( g \):
\begin{equation}
(u, o) = \mathcal{M}_{\text{LLM}}(s, s', \texttt{prompt}).
\end{equation}

\( u \) and \( o \) respectively correspond to the cost and benefit of successfully executing \( a \). Similarly, using LLMs, we can also obtain \( y = \mathcal{M}_{\text{LLM}}(s, s', \texttt{prompt'}) \), which represents the preconditions required for \( a \) to be successfully executed. More details in Algorithm \ref{alg:find_co} and \ref{alg:find_conditions}.

Next, we construct the sets \( \mathcal{U} \), \( \mathcal{O} \), and \( Y \). These sets aggregate the values of \( u \), \( o \), and \( y \) across various actions \( a \). Using these sets, we can form the experience \( \mathcal{E} \):

\begin{equation}
\mathcal{E} = \{ \mathcal{U}, \mathcal{O}, Y \}.
\end{equation}

In the "make wood pickaxe" action (objective), two preconditions must be met: "having at least 1 wood" and "standing next to a table". These preconditions serve as constraints, and the action given by policy is executed successfully only if both are satisfied. If not, the action results in a noop. Game laws define these preconditions as necessary, and we expect experience \(\mathcal{E}\) align with laws as closely as possible, though \(\mathcal{E}\) may sometimes be stricter, such as mistakenly requiring "at least 2 wood". Additionally, records indicating successful execution of the "make wood pickaxe" task also reflect its Costs and Benefits: consuming 1 wood to obtain 1 wood pickaxe. So for the action "make wood pickaxe", \( u \) is "\texttt{1 wood}", \( o \) is "\texttt{1 wood pickaxe}", and \( y \) is "\texttt{having at least 1 wood and standing next to a table}".

\subsection{LLMs Reasoning with $\mathcal{E}$}
LLMs have vast knowledge and can interact with environments through language~\cite{xi2023rise}. While they show zero-shot reasoning, it relies on prior knowledge and struggles in completely unfamiliar settings. Introducing \(\mathcal{E}\) helps address this limitation. By concatenating the original prompt \(p\) with \(\mathcal{E}\), \(p' = p \oplus \mathcal{E}\). This enhanced prompt \(p'\) is then used for reasoning by LLMs.

For instance, in \texttt{Minecraft}, LLMs can precisely guide how to craft a wood pickaxe by specifying that it requires 3 wood planks and 2 sticks. However, in similar other games, the recipe for crafting a wood pickaxe may vary. In such cases, agents can only attempt randomly, which is especially problematic without memory module, as each trial is isolated. By incorporating experience into reasoning, agents gain accurate knowledge about the environment.

\subsection{Reward Design with $\mathcal{E}$}
Carefully designed reward functions can help agents learn more quickly and perform better~\cite{li2017deep}. Previously, the design of reward functions often required expert involvement, now it can also be accomplished using LLMs~\cite{ma2023eureka, xie2023text2reward}. 
However, using LLMs to generate rewards primarily rely on feedback and iterative, making them difficult to apply in open-ended multi-task learning. Therefore, in contrast to these methods, our aim is to enable LLMs to optimize the structure of reward functions.

\begin{wrapfigure}[16]{r}{0.6\linewidth}
\begin{minipage}{0.6\textwidth}
\vspace{-22pt}
\begin{algorithm}[H]
\caption{Law-based Reward Generation with LLMs}
\label{alg:g_reward}
\begin{algorithmic}[1]
\small
\STATE \textbf{Require}: Objective set $\mathcal{G}$, environment experience $\mathcal{E}$, LLM model $\texttt{LLM}$, initial prompt $\texttt{prompt}$
\STATE \textbf{Hyperparameters}: Iterations $N$
\FOR{\textbf{each} $g \in \mathcal{G}$} 
    \STATE $r_g^{0} \gets \emptyset$ 
    \STATE \textcolor{gray}{// Extract relevant experience from $\mathcal{E}$ based on $g$}
    \STATE $e_g \gets \texttt{extract}(\mathcal{E}, g)$
    \FOR{\textbf{each} $i = 1$ to $N$}
        \STATE \textcolor{gray}{// $r_g^i$ is in code format}
        \STATE $r_g^i \sim \texttt{LLM}(g, e_g, \texttt{prompt} + r_g^{i-1})$ 
    \ENDFOR
    \STATE $r_g := r_g^{i}$ \quad \textcolor{gray}{// Reward function for objective $g$}
\ENDFOR
\STATE \textbf{Output}: $ \{ r_g \mid g \in G \} $ \quad \textcolor{gray}{// Set of reward functions}
\end{algorithmic}
\end{algorithm}
\end{minipage}
\end{wrapfigure}

Considering the experience \( e_i = \langle u_i, o_i, y_i \rangle \in \mathcal{E} \), which is related to the objective \( g_i \), where \( u_i \) records the element's cost to accomplish \( g_i \), \( o_i \) represents the expected outcome, and \( y_i \) shows the preconditions needed to achieve \( g_i \). The information contained within \( e_i \) approximately reflects the laws of the environment. Unlike methods that refine values~\cite{ma2023eureka}, we keep the reward values constant, but the determination of whether a reward is obtained at a particular step is controlled by generated code, as specified in Algorithm \ref{alg:g_reward}. \( e_i \) enhance the stability of code generation. LLMs only need to convert the conditions described in the text into code form and align it with the environment code. This law-based reward design method shifts the source of rewards from the environment to the agents themselves, acting as stepping stones that transform abstract knowledge into timely rewards.

\section{Experiments}

\texttt{Crafter} is a complex open-world game where agents must survive while continuously exploring the world. For more details about \texttt{Crafter}, please refer to Appendix \ref{appendix:cft}. We conducted experiments on different agents within \texttt{Crafter}: (1) LLM agents, where environment laws were directly incorporated into the context for the agents to reason about; (2) RL agents, where environment laws were translated into conditional judgment codes, yielding rewards. Implementation and experiments details can be found in Appendix \ref{appendix:detail}.

\textbf{LLM Agents.} We have adopted the framework proposed in SPRING~\cite{wu2024spring}. This framework not only allows agents to interact with the \texttt{Crafter} through text but also introduces a QA-DAG to help agents reason and make decisions more effectively. The QA-DAG, composed of 9 questions, is designed to promote a consistent chain-of-thought. SPRING traverses QA-DAG to determine the appropriate action for current state.
In SPRING, the agent configuration includes basic information about the environment and tasks, as well as specific details about \texttt{Crafter}, which is pre-extracted from the benchmark's \LaTeX{} source code~\cite{hafner2021benchmarking} via the "Paper Studying Module". 

\textbf{RL Agents.} We use \( \mathcal{E} \) to generate code for each achievement (or corresponding action / objective) to determine whether a reward should be granted, as described in Algorithm \ref{alg:g_reward}. Most achievements in \texttt{Crafter} correspond to a specific action, and an achievement is only accomplished when that action is performed under the right conditions. 

Specifically, we use the LLM in combination with \( \mathcal{E} \) to generate code for each action corresponding to an achievement, to evaluate the state when attempting an action. We use Proximal Policy Optimization (PPO)~\cite{schulman2017proximal} as our backbone algorithm and compare agents' performance under different rewards. Generated reward functions can be found in Appendix \ref{appendix:rcode}.

\begin{wraptable}[9]{r}{0.55\linewidth}
\vspace{-15pt}
\begin{center}
\begin{tabular}{l c c c}
\toprule
\multicolumn{1}{l}{Method} & \multicolumn{1}{c}{Score} & \multicolumn{1}{c}{Reward} \\
\toprule
SPRING + experience & $\bm{12.9\pm 2.3}$ & $\bm{9.9\pm 1.0}$ \\
SPRING + paper & $8.4\pm 1.4$ & $7.7\pm 0.8$ \\
SPRING + action & $2.0\pm 0.2$ & $1.3\pm 0.4$ \\
\bottomrule
\end{tabular}
\end{center}
\caption{\label{table:t1} Comparing SPRING with different context. More detailed context results in better agents. }
\vspace{-8mm}
\end{wraptable}

\textbf{Results.} We study the effect of different contextual information on the behavior of the LLM agent through ablations shown in Table \ref{table:t1}. 
We use the same reasoning framework from SPRING with different external content. "Experience" is text obtained using our method from records; "paper" represents information extracted from the original benchmark \LaTeX{}, consistent with the original SPRING implementation; "action" simply provides the names of the available actions. The basic game background is always provided.
Merely knowing the names of various actions results in significant hallucinations by the agent, similar to random attempts. Adding the key information extracted from the benchmark paper significantly improves performance. Using \( \mathcal{E} \) as context yields the best performance. We compared \( \mathcal{E} \) with the benchmark paper and found that the paper does not mention the recipes needed for crafting tools (e.g., what materials are needed to craft a stone pickaxe), which may affects the agent's expectations when selecting actions.

\begin{table}[H]
\vspace{-2mm}
\begin{center}
\begin{tabular}{l @{\hskip 0.50in} c c}
\toprule
\multicolumn{1}{l}{Method} & \multicolumn{1}{c}{Score} & \multicolumn{1}{c}{Training Steps} \\
\toprule
Human Experts & $50.5\pm 6.8\%$ & N/A \\
\midrule
health reward + achievement reward + penalty& $\bm{12.3 \pm 0.8\%}$ & 4M\\
health reward + achievement reward& $\phantom{0}9.7\pm 0.7\%$ & 4M\\
health reward & $\phantom{0}0.9\pm 0.1\%$ & 0\\
\bottomrule
\end{tabular}
\end{center}
\vspace{-1mm}
\caption{\label{table:t2} Compare the performance of agents trained under different rewards. After incorporating internal law-based rewards, agents improved.}
\vspace{-8mm}
\end{table}

In Table \ref{table:t2}, we study the performance of RL agents trained under different rewards. The most basic reward strategy only rewards the agent for changes in health, resulting in an agent that avoids deeper exploration (as it requires sacrificing health initially) and focuses on survival. Each action has a corresponding achievement reward generated by \( \mathcal{E} \). These functions assess whether an action is meaningful based on the agent's state and provide a 1-point reward for the first valid step of each action. The penalty is for poor choices: if the agent selects \( a \) in a state where none of \( a \)'s preconditions are met, it will incur -0.5 point penalty upon the first occurrence. The reward function actually only determines whether an action is valid (good enough) in a given state, and it can be combined with different reward shaping methods. The results show that law-based reward system with high-level reward shaping strategies can train better agents.

\section{Conclusions and Limitations}
The aim of this work is to leverage past interaction records to create experience, which can then be used to guide agents in future tasks. We propose a framework that converts interaction records from \texttt{Crafter} into textual experience, using these experience to develop LLM agents and train RL agents. Our findings show that these experience can approximate the laws governing the environment and provide internal motivation to assist agents in reasoning and reward design. By capturing the preconditions, costs, and benefits of each objective, agents can establish expectations about actions without relying on environmental feedback, enabling self-reflection and fostering self-motivation.

We propose a method for generating reward functions by determining the timing of rewards rather than iteratively adjusting their values. However, the rewards used in our experiments remain relatively simplistic. A potential avenue for improvement could involve combining experiential data with formal methods, such as using Finite State Machines (FSM), where reward machines~\cite{icarte2018using} can manage state transitions and reward design. The experience \(\mathcal{E}\) could approximate the environment's FSM and be transformed by LLMs to enhance integration with reward design methodologies.

\newpage

\bibliographystyle{plain}
\bibliography{neurips_2024}

\begin{thebibliography}{30}
\providecommand{\natexlab}[1]{#1}
\providecommand{\url}[1]{\texttt{#1}}
\expandafter\ifx\csname urlstyle\endcsname\relax
  \providecommand{\doi}[1]{doi: #1}\else
  \providecommand{\doi}{doi: \begingroup \urlstyle{rm}\Url}\fi

\bibitem[Altman(2021)]{altman2021constrained}
Eitan Altman.
\newblock \emph{Constrained Markov decision processes}.
\newblock Routledge, 2021.

\bibitem[Boud et~al.(2013)Boud, Keogh, and Walker]{boud2013reflection}
David Boud, Rosemary Keogh, and David Walker.
\newblock \emph{Reflection: Turning experience into learning}.
\newblock Routledge, 2013.

\bibitem[Dayan and Balleine(2002)]{dayan2002reward}
Peter Dayan and Bernard~W Balleine.
\newblock Reward, motivation, and reinforcement learning.
\newblock \emph{Neuron}, 36\penalty0 (2):\penalty0 285--298, 2002.

\bibitem[Du et~al.(2022)Du, Liu, Li, and Zhao]{du2022survey}
Yifan Du, Zikang Liu, Junyi Li, and Wayne~Xin Zhao.
\newblock A survey of vision-language pre-trained models.
\newblock \emph{arXiv preprint arXiv:2202.10936}, 2022.

\bibitem[Du et~al.(2023)Du, Watkins, Wang, Colas, Darrell, Abbeel, Gupta, and Andreas]{du2023guiding}
Yuqing Du, Olivia Watkins, Zihan Wang, C{\'e}dric Colas, Trevor Darrell, Pieter Abbeel, Abhishek Gupta, and Jacob Andreas.
\newblock Guiding pretraining in reinforcement learning with large language models.
\newblock In \emph{International Conference on Machine Learning}, pages 8657--8677. PMLR, 2023.

\bibitem[Dulac-Arnold et~al.(2019)Dulac-Arnold, Mankowitz, and Hester]{dulac2019challenges}
Gabriel Dulac-Arnold, Daniel Mankowitz, and Todd Hester.
\newblock Challenges of real-world reinforcement learning.
\newblock \emph{arXiv preprint arXiv:1904.12901}, 2019.

\bibitem[Fan et~al.(2022)Fan, Wang, Jiang, Mandlekar, Yang, Zhu, Tang, Huang, Zhu, and Anandkumar]{fan2022minedojo}
Linxi Fan, Guanzhi Wang, Yunfan Jiang, Ajay Mandlekar, Yuncong Yang, Haoyi Zhu, Andrew Tang, De-An Huang, Yuke Zhu, and Anima Anandkumar.
\newblock Minedojo: Building open-ended embodied agents with internet-scale knowledge.
\newblock \emph{Advances in Neural Information Processing Systems}, 35:\penalty0 18343--18362, 2022.

\bibitem[Griffith et~al.(2013)Griffith, Subramanian, Scholz, Isbell, and Thomaz]{griffith2013policy}
Shane Griffith, Kaushik Subramanian, Jonathan Scholz, Charles~L Isbell, and Andrea~L Thomaz.
\newblock Policy shaping: Integrating human feedback with reinforcement learning.
\newblock \emph{Advances in neural information processing systems}, 26, 2013.

\bibitem[Hafner(2021)]{hafner2021benchmarking}
Danijar Hafner.
\newblock Benchmarking the spectrum of agent capabilities.
\newblock \emph{arXiv preprint arXiv:2109.06780}, 2021.

\bibitem[Icarte et~al.(2018)Icarte, Klassen, Valenzano, and McIlraith]{icarte2018using}
Rodrigo~Toro Icarte, Toryn Klassen, Richard Valenzano, and Sheila McIlraith.
\newblock Using reward machines for high-level task specification and decomposition in reinforcement learning.
\newblock In \emph{International Conference on Machine Learning}, pages 2107--2116. PMLR, 2018.

\bibitem[Kojima et~al.(2022)Kojima, Gu, Reid, Matsuo, and Iwasawa]{kojima2022large}
Takeshi Kojima, Shixiang~Shane Gu, Machel Reid, Yutaka Matsuo, and Yusuke Iwasawa.
\newblock Large language models are zero-shot reasoners.
\newblock \emph{Advances in neural information processing systems}, 35:\penalty0 22199--22213, 2022.

\bibitem[Kwon et~al.(2023)Kwon, Xie, Bullard, and Sadigh]{kwon2023reward}
Minae Kwon, Sang~Michael Xie, Kalesha Bullard, and Dorsa Sadigh.
\newblock Reward design with language models.
\newblock \emph{arXiv preprint arXiv:2303.00001}, 2023.

\bibitem[Li(2017)]{li2017deep}
Yuxi Li.
\newblock Deep reinforcement learning: An overview.
\newblock \emph{arXiv preprint arXiv:1701.07274}, 2017.

\bibitem[Liu et~al.(2024)Liu, Feng, Wang, Wang, Liu, Zhao, Dengr, Ruan, Dai, Guo, et~al.]{liu2024deepseek}
Aixin Liu, Bei Feng, Bin Wang, Bingxuan Wang, Bo~Liu, Chenggang Zhao, Chengqi Dengr, Chong Ruan, Damai Dai, Daya Guo, et~al.
\newblock Deepseek-v2: A strong, economical, and efficient mixture-of-experts language model.
\newblock \emph{arXiv preprint arXiv:2405.04434}, 2024.

\bibitem[Ma et~al.(2023)Ma, Liang, Wang, Huang, Bastani, Jayaraman, Zhu, Fan, and Anandkumar]{ma2023eureka}
Yecheng~Jason Ma, William Liang, Guanzhi Wang, De-An Huang, Osbert Bastani, Dinesh Jayaraman, Yuke Zhu, Linxi Fan, and Anima Anandkumar.
\newblock Eureka: Human-level reward design via coding large language models.
\newblock \emph{arXiv preprint arXiv:2310.12931}, 2023.

\bibitem[Ng et~al.(1999)Ng, Harada, and Russell]{ng1999policy}
Andrew~Y Ng, Daishi Harada, and Stuart Russell.
\newblock Policy invariance under reward transformations: Theory and application to reward shaping.
\newblock In \emph{Icml}, volume~99, pages 278--287, 1999.

\bibitem[Park et~al.(2023)Park, O'Brien, Cai, Morris, Liang, and Bernstein]{park2023generative}
Joon~Sung Park, Joseph O'Brien, Carrie~Jun Cai, Meredith~Ringel Morris, Percy Liang, and Michael~S Bernstein.
\newblock Generative agents: Interactive simulacra of human behavior.
\newblock In \emph{Proceedings of the 36th annual acm symposium on user interface software and technology}, pages 1--22, 2023.

\bibitem[Schulman et~al.(2017)Schulman, Wolski, Dhariwal, Radford, and Klimov]{schulman2017proximal}
John Schulman, Filip Wolski, Prafulla Dhariwal, Alec Radford, and Oleg Klimov.
\newblock Proximal policy optimization algorithms.
\newblock \emph{arXiv preprint arXiv:1707.06347}, 2017.

\bibitem[Schutz(1962)]{schutz1962common}
Alfred Schutz.
\newblock Common-sense and scientific interpretation of human action.
\newblock In \emph{Collected papers I: The problem of social reality}, pages 3--47. Springer, 1962.

\bibitem[Skalse et~al.(2022)Skalse, Howe, Krasheninnikov, and Krueger]{skalse2022defining}
Joar Skalse, Nikolaus Howe, Dmitrii Krasheninnikov, and David Krueger.
\newblock Defining and characterizing reward gaming.
\newblock \emph{Advances in Neural Information Processing Systems}, 35:\penalty0 9460--9471, 2022.

\bibitem[Sun et~al.(2024)Sun, Zhuang, Kong, Dai, and Zhang]{sun2024adaplanner}
Haotian Sun, Yuchen Zhuang, Lingkai Kong, Bo~Dai, and Chao Zhang.
\newblock Adaplanner: Adaptive planning from feedback with language models.
\newblock \emph{Advances in Neural Information Processing Systems}, 36, 2024.

\bibitem[Wang et~al.(2023{\natexlab{a}})Wang, Xie, Jiang, Mandlekar, Xiao, Zhu, Fan, and Anandkumar]{wang2023voyager}
Guanzhi Wang, Yuqi Xie, Yunfan Jiang, Ajay Mandlekar, Chaowei Xiao, Yuke Zhu, Linxi Fan, and Anima Anandkumar.
\newblock Voyager: An open-ended embodied agent with large language models.
\newblock \emph{arXiv preprint arXiv:2305.16291}, 2023{\natexlab{a}}.

\bibitem[Wang et~al.(2023{\natexlab{b}})Wang, Lu, Santacroce, Gong, Zhang, and Shen]{wang2023adapting}
Kuan Wang, Yadong Lu, Michael Santacroce, Yeyun Gong, Chao Zhang, and Yelong Shen.
\newblock Adapting llm agents through communication.
\newblock \emph{arXiv preprint arXiv:2310.01444}, 2023{\natexlab{b}}.

\bibitem[Wu et~al.(2024{\natexlab{a}})Wu, Fan, Liang, Azaria, Li, and Mitchell]{wu2024read}
Yue Wu, Yewen Fan, Paul~Pu Liang, Amos Azaria, Yuanzhi Li, and Tom~M Mitchell.
\newblock Read and reap the rewards: Learning to play atari with the help of instruction manuals.
\newblock \emph{Advances in Neural Information Processing Systems}, 36, 2024{\natexlab{a}}.

\bibitem[Wu et~al.(2024{\natexlab{b}})Wu, Min, Prabhumoye, Bisk, Salakhutdinov, Azaria, Mitchell, and Li]{wu2024spring}
Yue Wu, So~Yeon Min, Shrimai Prabhumoye, Yonatan Bisk, Russ~R Salakhutdinov, Amos Azaria, Tom~M Mitchell, and Yuanzhi Li.
\newblock Spring: Studying papers and reasoning to play games.
\newblock \emph{Advances in Neural Information Processing Systems}, 36, 2024{\natexlab{b}}.

\bibitem[Xi et~al.(2023)Xi, Chen, Guo, He, Ding, Hong, Zhang, Wang, Jin, Zhou, et~al.]{xi2023rise}
Zhiheng Xi, Wenxiang Chen, Xin Guo, Wei He, Yiwen Ding, Boyang Hong, Ming Zhang, Junzhe Wang, Senjie Jin, Enyu Zhou, et~al.
\newblock The rise and potential of large language model based agents: A survey.
\newblock \emph{arXiv preprint arXiv:2309.07864}, 2023.

\bibitem[Xiang et~al.(2024)Xiang, Tao, Gu, Shu, Wang, Yang, and Hu]{xiang2024language}
Jiannan Xiang, Tianhua Tao, Yi~Gu, Tianmin Shu, Zirui Wang, Zichao Yang, and Zhiting Hu.
\newblock Language models meet world models: Embodied experiences enhance language models.
\newblock \emph{Advances in neural information processing systems}, 36, 2024.

\bibitem[Xie et~al.(2023)Xie, Zhao, Wu, Liu, Luo, Zhong, Yang, and Yu]{xie2023text2reward}
Tianbao Xie, Siheng Zhao, Chen~Henry Wu, Yitao Liu, Qian Luo, Victor Zhong, Yanchao Yang, and Tao Yu.
\newblock Text2reward: Automated dense reward function generation for reinforcement learning.
\newblock \emph{arXiv preprint arXiv:2309.11489}, 2023.

\bibitem[Zhang et~al.(2023)Zhang, Lehman, Stanley, and Clune]{zhang2023omni}
Jenny Zhang, Joel Lehman, Kenneth Stanley, and Jeff Clune.
\newblock Omni: Open-endedness via models of human notions of interestingness.
\newblock \emph{arXiv preprint arXiv:2306.01711}, 2023.

\bibitem[Zhu et~al.(2023)Zhu, Chen, Tian, Tao, Su, Yang, Huang, Li, Lu, Wang, et~al.]{zhu2023ghost}
Xizhou Zhu, Yuntao Chen, Hao Tian, Chenxin Tao, Weijie Su, Chenyu Yang, Gao Huang, Bin Li, Lewei Lu, Xiaogang Wang, et~al.
\newblock Ghost in the minecraft: Generally capable agents for open-world environments via large language models with text-based knowledge and memory.
\newblock \emph{arXiv preprint arXiv:2305.17144}, 2023.

\end{thebibliography}

\newpage
\appendix

\section{Related Works}
\label{appendix:related}
LLMs have a vast amount of knowledge, and although their overall performance is strong, additional context remains important for specific tasks. The most common approach is to place this context within the prompt, which is in-context learning~\cite{kojima2022large}. GITM~\cite{zhu2023ghost} improves agents' performance in games by introducing knowledge from the \texttt{Minecraft} Wiki, while SPRING~\cite{wu2024spring} allowing them to acquire information from academic papers. Generative Agents~\cite{park2023generative} reflect on and summarize the interaction history of agents, enabling believable simulations of human behavior.

In addition, researchers hope to leverage LLMs to benefit reinforcement learning agents. Designing reward functions using LLMs is an emerging approach~\cite{ma2023eureka, xie2023text2reward}. Despite having access to the environment's source code, existing LLMs often fail to generate optimal reward functions initially and heavily rely on iterative external feedback to incrementally refine these functions. 
Often, LLMs need to make subtle adjustments to hard-coded values in the code to achieve better design. 
Assuming there are, on average, $m$ combinations of values for a single task, then for $N$ tasks, there are $m^N$ possible combinations. Therefore, relying solely on feedback to adjust hard-coded values is impractical for multi-task learning. 
Moreover, this method also struggles with the challenges brought by reward hacking~\cite{skalse2022defining}.

\section{Starting from Records}
\label{appendix:records}
The objectives of agents need to be achieved under specific conditions. However, LLM or RL agents initially lack knowledge and understanding of the environment. Agents cannot determine which conditions are necessary and establish strategies only through trial-and-error. Successful records are worthy of attention. Agents are expected to benefit from these records when addressing the same objectives in the future. We propose using LLMs to reason through these records, extract useful information, and preserve it as experience.

Agents observe the environment in various forms, most commonly text and images. Existing Vision-Language Models (VLMs) are capable of describing images in natural language~\cite{du2022survey}, so we focus on extracting constraints from textual observations. Instead of merely structuring text descriptions, we expect LLMs to identify the constraints $\mathcal{C}$ of the objective, rather than documenting all conditions in the successful records, which is not conducive to generalization.

A record \( d = \langle s, a, s', v \rangle \) can be represented as a tuple, where \( s \) and \( s' \) are the states, \( s' \sim \mathcal{P}(\cdot \mid s, a) \), \( a \) is the action, and \( v \) is a boolean value indicating whether the agent successfully executed the action \( a \). 

As illustrated in Algorithm \ref{alg:find_conditions}, identifying necessary preconditions for objective \( g \) can be achieved by utilizing records from \( \mathcal{D}_g \). This method is specifically applicable to conditions connected by \textbf{AND}; however, for conditions connected by \textbf{OR} , they can be derived by leveraging more diverse \( \mathcal{D}_g \) and extracting from different records separately. Subsequently, LLMs can consolidate these findings at the language level. 

Additionally, enabling LLMs to infer the \texttt{why} and \texttt{when} of an action based on existing content can allow text-based agents to more explicitly access the relevant motivations.

\afterpage{
\begin{algorithm}[tb]
\setstretch{1.15}
\caption{Find Costs and Benefits for Objective $g$}
\label{alg:find_co}
\textbf{Require}: Language model \texttt{LLM}, extraction prompt $\texttt{prompt}_\texttt{e}$, update prompt $\texttt{prompt}_\texttt{u}$ \\
\textbf{Input}: Set \( \mathcal{D}_g^{v=\text{True}} \) \\
\textbf{Output}: String \( u_g \) and \( o_g \)
\begin{algorithmic}[1]
\STATE $u_g, o_g \gets \texttt{""}$
\FORALL{$d \in \mathcal{D}_g^{v=\text{True}}$}
    \STATE $x, x' \gets \texttt{LLM}(d, \texttt{prompt}_\texttt{e})$
    \STATE $\mathcal{U}_g, \mathcal{O}_g \gets \texttt{LLM}(u_g, o_g, x, x', \texttt{prompt}_\texttt{u})$
\ENDFOR
\STATE \textbf{return} \( u_g \), \( o_g \)
\end{algorithmic}
\end{algorithm}
}

\afterpage{
\begin{algorithm}[tb]
\setstretch{1.15}
\caption{Find Preconditions for Objective $g$}
\label{alg:find_conditions}
\textbf{Require}: Language model \texttt{LLM}, extraction prompt $\texttt{prompt}_\texttt{e}$, update prompt $\texttt{prompt}_\texttt{u}$ \\
\textbf{Input}: Sets \( \mathcal{D}_g \) and String \( u_g \) \\
\textbf{Output}: Sting \( y \), which is the necessary preconditions set for objective $g$
\begin{algorithmic}[1]
\STATE $y \gets \texttt{LLM}(u_g, \texttt{prompt}_\texttt{e})$
\FORALL{$d \in \mathcal{D}_g$}
    \STATE $Y \gets \texttt{LLM}(d, y, \texttt{prompt}_\texttt{u})$
\ENDFOR
\STATE \textbf{return} $y$
\end{algorithmic}
\end{algorithm}
}

\section{\texttt{Crafter} Details}
\label{appendix:cft}
\texttt{Crafter}~\cite{hafner2021benchmarking} is an abstraction of \texttt{Minecraft}, offering a simpler open-ended world.

\begin{figure*}[t]
\vspace*{-3ex}
\centering
\resizebox{0.9\linewidth}{!}{\includegraphics{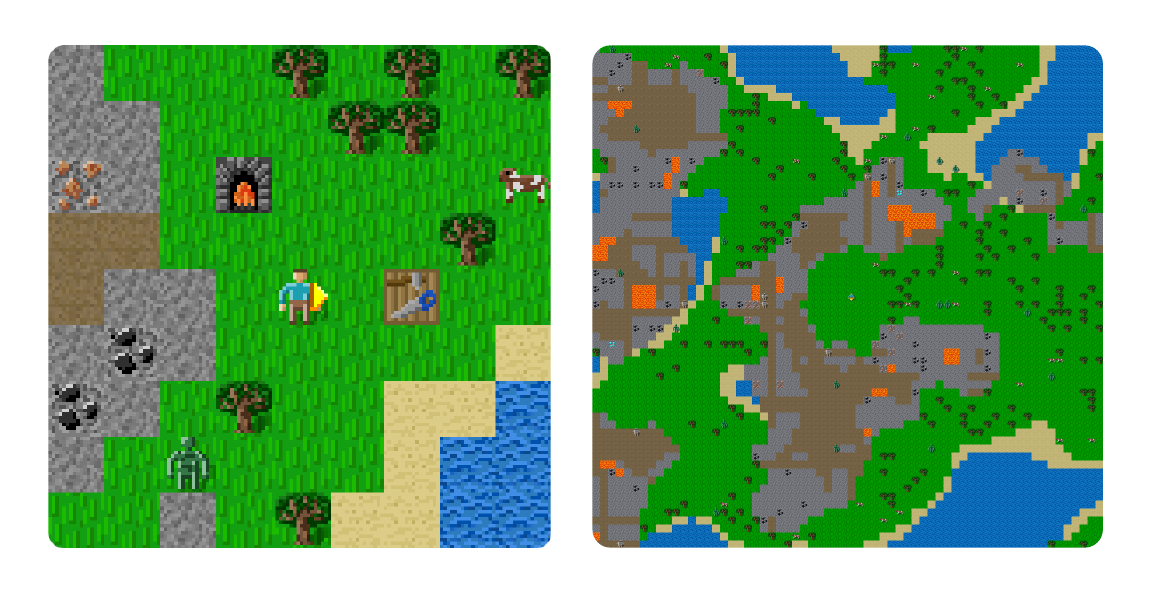}}
\caption{Each observation in \texttt{Crafter} is a $9 \times 9$ local map, and the entire world is generated by randomly combining various types of grids according to certain rules.}
\label{fig:wd}
\end{figure*}

\textbf{Game Mechanism.} Each game generates a world map randomly. Agents can collect resources and use them to craft items and tools, ensuring survival by maintaining health. The grid includes grass, water, stone, etc. Agents must keep their health, food, drink, and energy from dropping to zero. They can collect a variety of resources such as saplings, wood, coal, iron, and diamonds, etc, which can be used to craft tools or place on the grid. The positions and movements of other creatures are random, and killing these creatures provides additional resources.

\textbf{Interface.} The agent's observation is a frame of a $64 \times 64 \times 3$ image, displaying a map of the grid along with all items and information. The action space consists of 17 actions, some of which are only effective under specific conditions. A game ends when the player's health reach zero. For the training of RL agents, we referred to the modifications in ELLM~\cite{du2023guiding}, making the "do" (which means "attack" in front of a zombie but "drink" in front of water, has different effects in different situations) action more concrete by transforming it into specific actions such as "eat\_cow," "collect\_drink," "attack\_zombie," etc., resulting in a total of 11 actions. Consequently, the action space has been expanded to 27. This increases the exploration problem but also makes it compatible with our method.

\textbf{Challenges.} \texttt{Crafter} is able to assess various agent capabilities within a single environment, including survival, exploration, utilization, memory, and generalization. Due to the complexity of relationships between different tasks, agents need to explore deeply to accomplish more advanced tasks. To progress in the game, agents must repeatedly perform several actions over a long period, such as searching for food, defending against threats, and collecting materials that require multiple uses. More advanced agents should also remember position information to make decisions over extended periods. Lastly, because many game scenes are often similar in different environments, agents must recognize and remain robust against irrelevant changes across different environments.

\begin{figure*}[t]
\vspace*{-3ex}
\centering
\resizebox{0.7\linewidth}{!}{\includegraphics{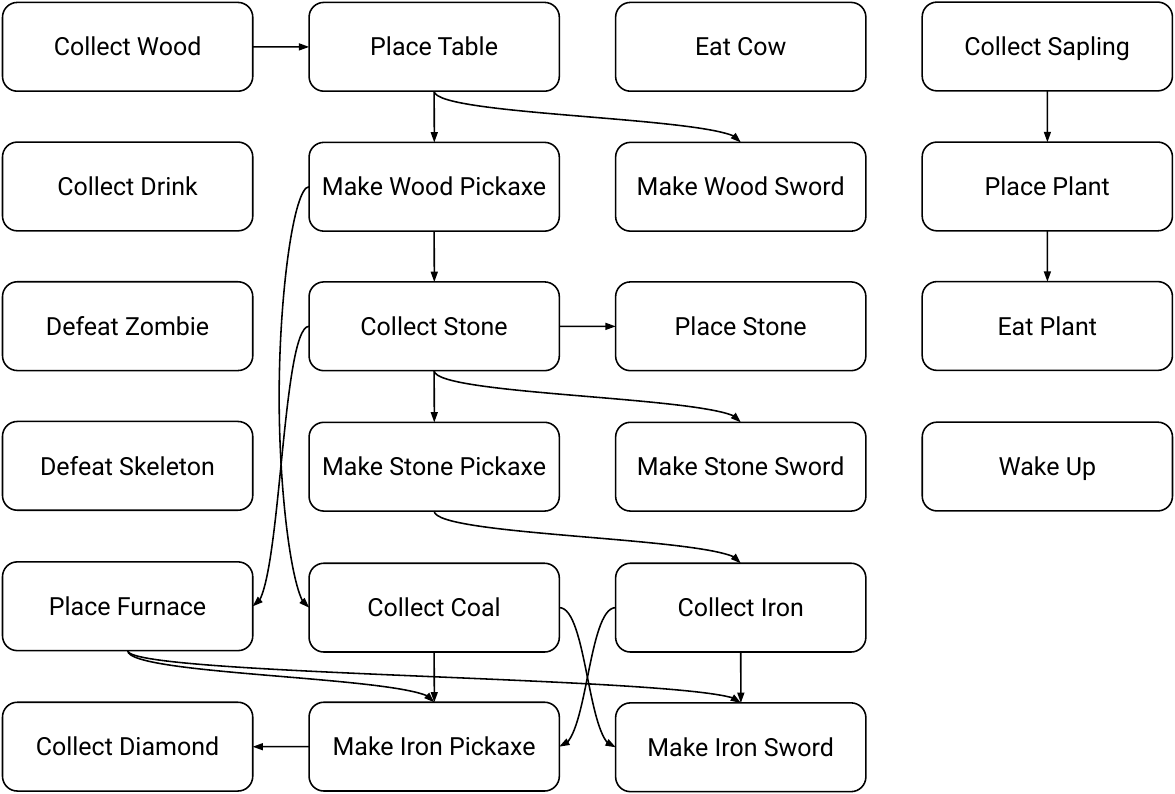}}
\caption{\texttt{Crafter} defines 22 achievements that can be unlocked in each episode of gameplay. The arrows represent the dependencies between achievements, indicating that the achievement being pointed to is one of the prerequisites for completing the achievement it points towards.}
\label{fig:achievements}
\end{figure*}

\textbf{Evaluation.} To evaluate various agent capabilities, \texttt{Crafter} defines 22 achievements, as illustrated in \ref{fig:achievements}. These achievements correspond to semantically meaningful behaviors, such as gathering various resources, crafting items and tools, seeking food and water, defeating monsters, and more. These achievements encompass a range of difficulty levels, with some being independent of each other, thus testing the breadth of exploration, while others are interdependent, testing the depth of exploration.

Based on these 22 achievements, agents can be evaluated through \textbf{Success rate} and \textbf{Score}.

\begin{itemize}
\item \textbf{Success Rate} \quad Each achievement corresponds to a success rate $R_i$, which is defined as the ratio of the number of episodes where the agent successfully completes the achievement to the total number of episodes. Here, $N_i$ represents the number of episodes in which achievement $i$ is accomplished, and $N_{total}$ represents the total number of episodes.
    \begin{equation}
        R_i = \frac{N_i}{N_{total}}
    \end{equation}
\item \textbf{Score} \quad Score is calculated by aggregating the success rates of individual achievements. Unlocking high-difficulty achievements, even if infrequent, should be more significant than further increasing the success rate of already unlocked achievements. Here, \(s_i \in [0, 100]\) represents the agent’s success rate for achievement $i$, and \(N = 22\) is the total number of achievements.
    \begin{equation}
        S = \exp\left(\frac{1}{N}\sum_{i=1}^N \ln\left(1 + s_i\right)\right) - 1
    \end{equation}
\end{itemize}

\section{Implementation Details}
\label{appendix:detail}

For records used to infer experience, we represent them as a triplet composed of "the description of the state, the action, and the description of the state after the action is performed." An example is as follows. The reason for choosing \texttt{attributes}, \texttt{tools}, \texttt{materials}, \texttt{face}, and \texttt{nearby} to describe a state is determined by \texttt{Crafter}.

In all parts of the experiment involving LLM access, we used DeepSeek-V2~\cite{liu2024deepseek} from DeepSeek's API.

\subsection{Data Collection}
\label{appendix:data}
Before starting, we collected structured interaction records from human players. In \texttt{Crafter}, 22 predefined achievements exist, each of which requires performing a specific action under particular conditions. We gathered 10 successful and 10 failed attempts for each achievement and summarized the experience \( \mathcal{E} \) using the method described in Algorithm \ref{alg:find_co} and \ref{alg:find_conditions}. Under appropriate prompt settings and with the interaction records, LLMs were able to generate \( \mathcal{E} \) that aligned with the environment laws. However, we found that the diversity of records had a significant impact on the LLM's reasoning capabilities. Additionally, we observed that LLMs appeared capable of determining whether an attempt was successful or failed by comparing the states before and after, without needing explicit labels. All the prompts can be found in Appendix \ref{appendix:prompt}.

\subsection{Interaction Records}
\label{appendix:interactionlb}

\newpage
\begin{tcolorbox}[breakable=true, boxrule={0.5pt}, sharp corners={all}]
\setlength{\parskip}{1ex}
action: collect\_wood \\

init\_state: \\
\hbox{\ \ }\{ \\
\hbox{\ \ \ \ }"attributes": \{ \\
\hbox{\ \ \ \ \ \ \ }"health": 9, \\
\hbox{\ \ \ \ \ \ \ }"food": 8, \\
\hbox{\ \ \ \ \ \ \ }"drink": 7, \\
\hbox{\ \ \ \ \ \ \ }"energy": 7 \\
\hbox{\ \ \ \ }\}, \\
\hbox{\ \ \ \ }"tools": \{\}, \\
\hbox{\ \ \ \ }"materials": \{ \\
\hbox{\ \ \ \ \ \ \ }"sapling": 1 \\
\hbox{\ \ \ \ }\}, \\
\hbox{\ \ \ \ }"face": "tree()", \\
\hbox{\ \ \ \ }"nearby": [ \\
\hbox{\ \ \ \ \ \ \ }["grass()", "tree()", "grass()"], \\
\hbox{\ \ \ \ \ \ \ }["grass()", "grass(player)", "grass()"], \\
\hbox{\ \ \ \ \ \ \ }["grass(cow)", "grass()", "grass()"] \\
\hbox{\ \ \ \ }] \\
\hbox{\ \ } \} \\

resulting\_state: \\
\hbox{\ \ }\{ \\
\hbox{\ \ \ \ }"attributes": \{ \\
\hbox{\ \ \ \ \ \ \ }"health": 9, \\
\hbox{\ \ \ \ \ \ \ }"food": 8, \\
\hbox{\ \ \ \ \ \ \ }"drink": 7, \\
\hbox{\ \ \ \ \ \ \ }"energy": 7 \\
\hbox{\ \ \ \ }\}, \\
\hbox{\ \ \ \ }"tools": \{\}, \\
\hbox{\ \ \ \ }"materials": \{ \\
\hbox{\ \ \ \ \ \ \ }"wood": 1, \\
\hbox{\ \ \ \ \ \ \ }"sapling": 1 \\
\hbox{\ \ \ \ }\}, \\
\hbox{\ \ \ \ }"face": "grass()", \\
\hbox{\ \ \ \ }"nearby": [ \\
\hbox{\ \ \ \ \ \ \ }["grass()", "grass()", "grass()"], \\
\hbox{\ \ \ \ \ \ \ }["grass()", "grass(player)", "grass()"], \\
\hbox{\ \ \ \ \ \ \ }["grass(cow)", "grass()", "grass()"] \\
\hbox{\ \ \ \ }] \\
\hbox{\ \ } \} \\

\end{tcolorbox}

The interaction between the LLM agent and \texttt{Crafter} is conducted through text. SPRING~\cite{wu2024spring} has implemented text-based interaction for \texttt{Crafter}. Here is an example.

\begin{tcolorbox}[breakable=true, boxrule={0.5pt}, sharp corners={all}]
\setlength{\parskip}{1ex}

You see: \\
\hbox{\ \ }- grass 1 steps to your south \\
\hbox{\ \ }- stone 3 steps to your west \\
\hbox{\ \ }- path 1 steps to your west \\ 
\hbox{\ \ }- tree 5 steps to your north-east \\
\hbox{\ \ }- table 1 steps to your east \\
\hbox{\ \ }- furnace 2 steps to your south-west \\
\hbox{\ \ }- zombie 2 steps to your south-east \\
\hbox{\ \ }- plant 1 steps to your north \\

You face plant at your front.\\

Your attributes status: \\
\hbox{\ \ }- health: 2/9 \\
\hbox{\ \ }- food: 2/9 \\
\hbox{\ \ }- drink: 0/9 \\
\hbox{\ \ }- energy: 4/9 \\

Your materials inventory: \\
\hbox{\ \ }- stone: 1 \\
\hbox{\ \ }- iron: 5 \\

Your tools inventory: \\
\hbox{\ \ }- wood\_pickaxe: 2 \\
\hbox{\ \ }- stone\_pickaxe: 4 \\
\hbox{\ \ }- stone\_sword: 1"

\end{tcolorbox}

\subsection{Prompt}
\label{appendix:prompt}
Here is the prompt used for extracting experience from interaction records: \texttt{SYSTEM + USER A} is used to generate results based solely on the common knowledge of the LLM itself, while \texttt{SYSTEM + existing experience + USER B} is used for refining the output.

\begin{tcolorbox}[breakable=true, boxrule={0.5pt}, sharp corners={all}]
\setlength{\parskip}{1ex}
\textbf{SYSTEM:}\\
You are a player who is in an open-world game. It's up to you to explore as much of the world while trying to survive! The world is made of grids.

ATTRIBUTES are some information related to yourself. These are the attributes that you have to manage in order to survive. They are affected by the your actions and the environment. All max values are 9.\\
\{\\
\hbox{\ \ }"health", \\
\hbox{\ \ }"food", \\
\hbox{\ \ }"drink", \\
\hbox{\ \ }"energy"\\
\}\\

TOOLS are some of the tools you currently have. In accomplishing some actions, you can use the tools that are held, and you can also make more tools.

MATERIALS are some materials you currently have, which can be obtained by interacting with the environment. You can combine them into a construction tool, or for other purposes.

FACE records the grid you're currently facing. In some special cases, the object on the current grid is recorded in '()'.

NEARBY records a nine-panel grid centered on you, the player. \\

\end{tcolorbox}
\begin{tcolorbox}[breakable=true, boxrule={0.5pt}, sharp corners={all}]
\setlength{\parskip}{1ex}

\textbf{USER A:}\\
Let's consider an action called "\{action\_name\}", what kind of things do you think this action does? Do you guess what the effect would be on some element in "\{aspect\}"? You only need to considering the changes in "\{aspect\}".

Completing an action costs something (optional) and gains some benefit (optional). Please pay attention to the difference between "initial\_state" and "resulting\_state". Describe in natural language what happened in this transition.

You only need to output one description without any other words. \\
\end{tcolorbox}

\begin{tcolorbox}[breakable=true, boxrule={0.5pt}, sharp corners={all}]
\setlength{\parskip}{1ex}
\textbf{USER B:} \\
Now, I will show you the comparison of "\{aspect\}" before and after the player executes action "\{action\_name\}". \\

\{records\}\\

Please pay attention to the difference between "initial\_state" and "resulting\_state". 

\end{tcolorbox}

After obtaining the experience, use the following prompt, along with partial environment code, to generate code that determines whether each action is valid.

\begin{tcolorbox}[breakable=true, boxrule={0.5pt}, sharp corners={all}]
\setlength{\parskip}{1ex}
\textbf{SYSTEM:}\\
The following information can help you in the process of designing a Reward Function:

The game environment is consist of a grid of blocks. Each block has a texture and an object on it (optional). The texture can be one of the following:\\
\hbox{\ \ }- water \\
\hbox{\ \ }- grass \\
\hbox{\ \ }- stone \\
\hbox{\ \ }- path \\
\hbox{\ \ }- sand \\
\hbox{\ \ }- tree \\
\hbox{\ \ }- lava \\
\hbox{\ \ }- coal \\
\hbox{\ \ }- iron \\
\hbox{\ \ }- diamond \\
\hbox{\ \ }- table \\
\hbox{\ \ }- furnace \\

The objects can be:\\
\hbox{\ \ }- Zombie \\
\hbox{\ \ }- Skeleton \\
\hbox{\ \ }- Plant \\
\hbox{\ \ }- Cow \\

Agent can perform the following actions:\\
\hbox{\ \ }- noop \\
\hbox{\ \ }- move\_left \\
\hbox{\ \ }- move\_right \\
\hbox{\ \ }- move\_up \\
\hbox{\ \ }- move\_down \\
\hbox{\ \ }- eat\_plant \\
\hbox{\ \ }- defeat\_zombie \\
\hbox{\ \ }- defeat\_skeleton \\
\hbox{\ \ }- eat\_cow \\
\hbox{\ \ }- collect\_coal \\
\hbox{\ \ }- collect\_diamond \\
\hbox{\ \ }- collect\_drink \\
\hbox{\ \ }- collect\_iron \\
\hbox{\ \ }- collect\_sapling \\
\hbox{\ \ }- collect\_stone \\
\hbox{\ \ }- collect\_wood \\
\hbox{\ \ }- sleep \\
\hbox{\ \ }- place\_stone \\
\hbox{\ \ }- place\_table \\
\hbox{\ \ }- place\_furnace \\
\hbox{\ \ }- place\_plant \\
\hbox{\ \ }- make\_wood\_pickaxe \\
\hbox{\ \ }- make\_stone\_pickaxe \\
\hbox{\ \ }- make\_iron\_pickaxe \\
\hbox{\ \ }- make\_wood\_sword \\
\hbox{\ \ }- make\_stone\_sword \\
\hbox{\ \ }- make\_iron\_sword \\

\textbf{ATTRIBUTES} are some information related to the agent. These are the attributes that the agent have to manage in order to survive. They are affected by the agent's actions and the environment. All max values are 9.
\begin{verbatim}
{
    "health", 
    "food", 
    "drink", 
    "energy"
}
\end{verbatim}

\textbf{TOOLS} are some of the tools the agent currently have. In accomplishing some actions, the agent can use the tools that are held, and the agent can also make more tools.

\textbf{MATERIALS} are some materials the agent currently have, which can be obtained by interacting with the environment. Agent can combine them into a construction tool, or for other purposes.

\textbf{FACE} records the grid the agent is currently facing. 

\textbf{NEARBY} records a nine-panel grid centered on the agent.

When you help AGENT with Reward Function Design, you may also need some code-level knowledge, which can help you better translate your understanding into sensible Python code:

- You can visit the AGENT's inventory by calling the function agent.inventory. 
  It will return a dictionary with the resources and tools that the AGENT has. 
  agent.inventory including information of ATTRIBUTES, TOOLS and MATERIALS. \\
e.g., \\
\texttt{agent.inventory \\
\# \{'health': 9, 'food': 9, 'drink': 9, 'energy': 9, 'sapling': 0, 'wood': 0, 
   'stone': 0, 'coal': 0, 'iron': 0, 'diamond': 0, 'wood\_pickaxe': 0, 
   'stone\_pickaxe': 0, 'iron\_pickaxe': 0, 'wood\_sword': 0, 'stone\_sword': 0, 
   'iron\_sword': 0\}} \\
\\
\\
- You can get information about the gird the agent is facing by accessing 
  agent.world[target]. Gird is probably some kind of texture or an object.\\
e.g.,\\
\texttt{texture, obj = agent.world[target]\\
\# texture is a string and obj is a object. Cow, Zombie, Skeleton, Plant, are a 
  list of objects and others are all texture. treat objects using isinstance().}\\
\\
\\
- You can look at the NEARBY AGENT by agent.world.nearby(agent.pos, 1). 
  Similar to facing, this function call will return a 'tuple', which contains a 
  tuple of materials (string), and a set of objects.\\
e.g.,
\texttt{agent.world.nearby(agent.pos, 1)\\
\# (('grass', 'sand'), \{\{<crafter.objects.Plant object at 0x7f4283106290>, 
   <crafter.objects.Zombie object at 0x7f4283106440>, <crafter.objects.Player 
   object at 0x7f42845c9960>\}\})}

\end{tcolorbox}

\begin{tcolorbox}[breakable=true, boxrule={0.5pt}, sharp corners={all}]
\setlength{\parskip}{1ex}
\textbf{USER:} \\
Now you need to write a reward function, which is a simple function that only needs to determine if the action can be done in the current state, the action is: \{action\_name\}

Here are some understanding of this action: \\
\{experience\} \\

You only need to output the python function named '\{name\}\_reward(agent, target)' and the function return a bool value.\\
'True' means the action can be done at current state, 'False' means the action can not be done at current state.\\

Output code only, without any explanation.
\end{tcolorbox}

\subsection{Experience in Text}
\textbf{Preconditions}
\begin{tcolorbox}[breakable=true, boxrule={0.5pt}, sharp corners={all}]
\setlength{\parskip}{1ex}
1. Collect Wood: collected wood from a tree, adding it to the player's inventory while leaving the player's attributes and tools unchanged, and removing the tree from the grid the player was facing. \\
2. Place Table: consumes 2 units of wood to place a table in the player's facing grid, replacing grass, without affecting attributes, tools, or nearby entities, potentially offering new interaction options. \\
3. Eat Cow: increases the player's 'food' attribute by 6 points, removes the cow from the facing grid, and has no other observed effects on attributes, tools, materials, or the nearby environment. \\
4. Collect Sapling: added one sapling, had no effect on ATTRIBUTES or TOOLS, did not change the FACE attribute. \\
5. Collect Drink: increased the player's 'drink' attribute by 1 without affecting other attributes, tools, materials, or the environment, indicating a focused hydration replenishment with no visible environmental impact. \\
6. Make Wood Pickaxe: consumes 1 unit of wood, adds a wood pickaxe to the player's tools, and leaves all attributes and the environment unchanged. \\
7. Make Wood Sword: consumes one unit of 'wood' from the player's materials, adds a 'wood\_sword' to their tools, and does not affect attributes or the environment. \\
8. Place Plant: consumed a sapling, added a plant to the player's current grid, and had no immediate impact on attributes or tools. \\
9. Defeat Zombie: resulted in the removal of a zombie from the grid in front of the player, with no changes to the player's attributes, tools, materials, or the environment. \\
10. Collect Stone: collects a stone from the player's facing grid, adds it to the inventory, and reveals a path in the now-empty grid, with no impact on attributes or tools. \\
11. Place Stone: transitions the player's facing grid to stone, reducing the stone inventory by one without affecting attributes, tools, or other materials. \\
12. Eat Plant: increases the player's 'food' attribute by 4 and removes a ripe plant from the facing grid, with no other attribute or environmental changes. \\
13. Defeat Skeleton: resulted in the removal of the skeleton from the player's facing grid without affecting attributes, tools, or materials, indicating a neutral combat encounter with no immediate rewards or resource changes. \\
14. Make Stone Pickaxe: consumes 1 unit of wood and 1 unit of stone to craft a stone pickaxe, leaving the player's attributes unchanged and not affecting the nearby environment. \\
15. Make Stone Sword: successfully crafts a stone sword using 1 stone and 1 wood from the player's inventory, without affecting the player's attributes or the environment. \\
16. Sleep: transitions the player's state to "sleeping" without altering attributes, tools, materials, or the environment, suggesting a focus on internal attribute restoration without external impact. \\
17. Place Furnace: consumes 4 stones to place a furnace on the grid the player is facing, without affecting the player's attributes, tools, or position, and without immediate environmental impact. \\
18. Collect Coal: Removed coal from the grid the player was facing, added it to the player's materials inventory, and replaced the coal grid with a path, without affecting the player's attributes or tools. \\
19. Collect Iron: requires stone\_pickaxe and facing iron. Successfully collected iron from the grid directly to the right of the player, converting it from an iron() to path(), without affecting other attributes or nearby grids. \\
20. Make Iron Pickaxe: requires 1 wood, 1 coal, and 1 iron, and table and furnace nearby. Has resulted in the player crafting an iron pickaxe, consuming 1 unit of wood, 1 unit of iron, and coal, while adding the iron pickaxe to their tools without affecting their attributes or the environment. \\
21. Make Iron Sword: requires 1 wood, 1 coal, and 1 iron, and table and furnace nearby. Successfully crafts an iron sword, consuming one iron and one wood from the player's materials, without affecting health, food, drink, or energy, and leaves the environment unchanged. \\
22. Collect Diamond: requires iron\_pickaxe and facing diamond. Successfully adds a diamond to the player's materials inventory while transforming the faced grid from a diamond-containing area to a path, with no impact on attributes, tools, or other nearby resources.
\end{tcolorbox}

\textbf{Costs \& Benefits}
\begin{tcolorbox}[breakable=true, boxrule={0.5pt}, sharp corners={all}]
\setlength{\parskip}{1ex}
1. Collect Wood: Requires facing tree \\
2. Place Table: Requires 2 woods and facing grass or sand or path \\
3. Eat Cow: Requires facing a cow \\
4. Collect Sapling: Requires facing grass \\
5. Collect Drink: Requires facing water \\
6. Make Wood Pickaxe: Requires 1 wood and table nearby \\
7. Make Wood Sword: Requires 1 wood and table nearby \\
8. Place Plant: Requires 1 sapling and facing grass \\
9. Defeat Zombie: Requires facing zombie and better with weapons \\
10. Collect Stone: Requires wood\_pickaxe and facing stone \\
11. Place Stone: Requires 1 stone and facing grass or sand or path or water or lava \\
12. Eat Plant: Requires facing ripe plant \\
13. Defeat Skeleton: Requires facing skeleton and better with weapons \\
14. Make Stone Pickaxe: Requires 1 wood and 1 stone and table nearby \\
15. Make Stone Sword: Requires 1 wood and 1 stone and table nearby \\
16. Sleep: Requires insufficient energy \\
17. Place Furnace: Requires 4 stones and facing grass or sand or path \\
18. Collect Coal: Requires wood\_pickaxe and facing coal \\
19. Collect Iron: Requires stone\_pickaxe and facing iron \\
20. Make Iron Pickaxe: Requires 1 wood and 1 coal and 1 iron, also need table and furnace nearby \\
21. Make Iron Sword: Requires 1 wood and 1 coal and 1 iron, also need table and furnace nearby \\
22. Collect Diamond: Requires iron\_pickaxe and facing diamond
\end{tcolorbox}

\subsection{Reward Functions}
\label{appendix:rcode}
\begin{tcolorbox}[breakable=true, boxrule={0.5pt}, sharp corners={all}]
\setlength{\parskip}{1ex}
\begin{lstlisting}[style=python]
def collect_coal_reward(agent, target):
    texture, obj = agent.world[target]
    if texture == 'coal' and agent.inventory['wood_pickaxe'] > 0:
        return True
    return False

def eat_plant_reward(agent, target):
    texture, obj = agent.world[target]
    return isinstance(obj, Plant)

def defeat_zombie_reward(agent, target):
    texture, obj = agent.world[target]
    if isinstance(obj, Zombie):
        if 'iron_sword' in agent.inventory or 'stone_sword' in agent.inventory or 'wood_sword' in agent.inventory:
            return True
    return False

def defeat_skeleton_reward(agent, target):
    texture, obj = agent.world[target]
    if isinstance(obj, Skeleton):
        if agent.inventory['wood_sword'] > 0 or agent.inventory['stone_sword'] > 0 or agent.inventory['iron_sword'] > 0:
            return True
    return False

def eat_cow_reward(agent, target):
    texture, obj = agent.world[target]
    return isinstance(obj, Cow)

def collect_coal_reward(agent, target):
    texture, obj = agent.world[target]
    if texture == 'coal' and agent.inventory['wood_pickaxe'] > 0:
        return True
    return False

def collect_diamond_reward(agent, target):
    texture, obj = agent.world[target]
    if texture == 'diamond' and agent.inventory['iron_pickaxe'] > 0:
        return True
    return False

def collect_drink_reward(agent, target):
    texture, obj = agent.world[target]
    return texture == 'water'

def collect_iron_reward(agent, target):
    texture, obj = agent.world[target]
    return texture == 'iron' and 'stone_pickaxe' in agent.inventory

def collect_sapling_reward(agent, target):
    texture, obj = agent.world[target]
    return texture == 'grass'

def collect_stone_reward(agent, target):
    texture, obj = agent.world[target]
    return texture == 'stone' and 'wood_pickaxe' in agent.inventory

def collect_wood_reward(agent, target):
    texture, obj = agent.world[target]
    return texture == 'tree'

def sleep_reward(agent, target):
    return agent.inventory['energy'] < 9

def place_stone_reward(agent, target):
    if agent.inventory['stone'] < 1:
        return False
    texture, obj = agent.world[target]
    if texture not in ['grass', 'sand', 'path', 'water', 'lava']:
        return False
    return True

def place_table_reward(agent, target):
    texture, obj = agent.world[target]
    if texture in ['grass', 'sand', 'path'] and 'wood' in agent.inventory and agent.inventory['wood'] >= 2:
        return True
    return False

def place_furnace_reward(agent, target):
    if agent.inventory['stone'] >= 4:
        texture, _ = agent.world[target]
        if texture in ['grass', 'sand', 'path']:
            return True
    return False

def place_plant_reward(agent, target):
    if agent.inventory['sapling'] >= 1 and agent.world[target][0] == 'grass':
        return True
    return False

def make_wood_pickaxe_reward(agent, target):
    if agent.inventory['wood'] >= 1 and any(isinstance(obj, Table) for obj in agent.world.nearby(agent.pos, 1)[1]):
        return True
    return False

def make_stone_pickaxe_reward(agent, target):
    if 'wood' in agent.inventory and 'stone' in agent.inventory and 'table' in agent.world[target][1]:
        return True
    return False

def make_iron_pickaxe_reward(agent, target):
    materials = agent.inventory
    if materials['wood'] < 1 or materials['coal'] < 1 or materials['iron'] < 1:
        return False
    nearby = agent.world.nearby(agent.pos, 1)
    if 'table' not in nearby[0] or 'furnace' not in nearby[0]:
        return False
    return True

def make_wood_sword_reward(agent, target):
    if agent.inventory['wood'] >= 1:
        nearby_textures, nearby_objects = agent.world.nearby(agent.pos, 1)
        if 'table' in nearby_textures:
            return True
    return False

def make_stone_sword_reward(agent, target):
    inventory = agent.inventory
    if inventory['wood'] >= 1 and inventory['stone'] >= 1:
        nearby = agent.world.nearby(agent.pos, 1)
        for texture, obj in nearby:
            if 'table' in texture:
                return True
    return False

def make_iron_sword_reward(agent, target):
    inventory = agent.inventory
    if inventory['wood'] >= 1 and inventory['coal'] >= 1 and inventory['iron'] >= 1:
        nearby = agent.world.nearby(agent.pos, 1)
        textures, objects = nearby
        if 'table' in textures and 'furnace' in textures:
            return True
    return False
\end{lstlisting}
\end{tcolorbox}

\end{document}